\documentclass[10pt,journal,compsoc]{IEEEtran}
\ifCLASSOPTIONcompsoc
  \usepackage[nocompress]{cite}
\else
  \usepackage{cite}
\fi
\ifCLASSINFOpdf
\else
\fi
\usepackage{amsmath}
\usepackage{multirow}
\usepackage[utf8]{inputenc} 
\usepackage[T1]{fontenc}    
\usepackage{hyperref}       
\usepackage{url}            
\usepackage{booktabs}       
\usepackage{amsfonts}       
\usepackage{nicefrac}       
\usepackage{microtype}      
\usepackage[pdftex]{graphicx}
\usepackage	{subfig}
\usepackage{spverbatim}

\begin{document}
%
\title{Neural Networks for Text Correction and Completion in Keyboard Decoding}
%
%
%
%
\author{Shaona~Ghosh
        and~Per~Ola~Kristensson}
\markboth{Journal of \LaTeX\ Class Files,~Vol.~14, No.~8, August~2015}%
{Shell \MakeLowercase{\textit{et al.}}: Bare Demo of IEEEtran.cls for Computer Society Journals}
%



\IEEEtitleabstractindextext{%
\begin{abstract}
Despite the ubiquity of mobile and wearable text messaging applications, the problem of keyboard text decoding is not tackled sufficiently in the light of the enormous success of the deep learning Recurrent Neural Network (RNN) and Convolutional Neural Networks (CNN) for natural language understanding. In particular, considering that the keyboard decoders should operate on devices with  memory and processor resource constraints, makes it challenging to deploy industrial scale deep neural network (DNN) models. This paper proposes a sequence-to-sequence neural attention network system for automatic text correction and completion. Given an erroneous sequence, our model encodes character level hidden representations and then decodes the revised sequence thus enabling auto-correction and completion. 
Further, what makes the problem different from vanilla language modelling is the simultaneous text correction and completion.  We achieve this by a combination of character level CNN and gated recurrent unit (GRU) encoder along with and a word level gated recurrent unit (GRU) attention decoder. Unlike traditional language models that learn from billions of words, our corpus size is only $12$ million words; an order of magnitude smaller. The memory footprint of our learnt model for inference and prediction is also an order of magnitude smaller than the conventional language model based text decoders. We report baseline performance for neural keyboard decoders in such limited domain.
Our models achieve a word level accuracy of $90\%$ and a character error rate {\tt CER} of $2.4$ over the Twitter typo dataset. We present a novel dataset of noisy to corrected mappings by inducing the noise distribution from the Twitter data over the OpenSubtitles $2009$ dataset; on which our model predicts with a word level accuracy of $98\%$ and sequence accuracy of $0.689$. We have also conducted an user study from $8$ users, with our model predicting with an average {\tt CER} of $2.6\%$ while being competitive with the state-of-the-art non-neural touch-screen keyboard decoders at {\tt CER} of $1.6\%$. We observe a {\tt CER} of $2.1\%$ on physical keyboard based decoding. Further, we plan to release the training dataset and the associated software along with the baselines as an open-source tool-kit. We also propose an alternative smooth evaluation measure over the character error rate (CER) for evaluating model predictions based on the contextual underlying intent of the sentence.  Additionally, we have released our trained decoder as an inference server available at \url{www-edc.eng.cam.ac.uk/shaona}.
\end{abstract}
}

\maketitle
\IEEEdisplaynontitleabstractindextext
%
\IEEEpeerreviewmaketitle
\IEEEraisesectionheading{\section{Introduction}\label{sec:introduction}}
With the ubiquity of touch-screen mobile text interfaces, simultaneous error correction and completion is an important problem that cannot be solved using stand-alone language models without the availability of dense repetitive error patterns in the training data. Error patterns vary with user's style of typing, interface design, text entry method, user's native language, character and vocabulary restrictions among others. Traditionally, text correction and completion systems are integrated as keyboard decoders, and use a keyboard likelihood model combined with prior probability from the language model~\cite{Vertanen, kristensson2004shark, zhai2002movement, zhai2008interlaced, Vertanen}. Such models assign probability over words or characters and estimate the likelihood of the current word or character based on the previous words or characters in the sequence~\cite{Heafield-estimate, Heafield-kenlm}. 

In contrast, in the recent deep learning literature, recurrent neural network (RNN)~\cite{rodriguez1999recurrent} architectures have achieved success on a wide variety of sequence problems. They are extremely efficient to model the underlying natural language. In fact, recurrent neural networks, long short-term memory networks~\cite{hochreiter1997long} and gated recurrent neural networks~\cite{chung2014empirical}  have become standard approaches in sequence modelling and transduction problems such as language modelling and machine translation~\cite{sutskever2014sequence, bahdanau2014neural, cho2014learning}. 
The very recent model architecture called Transformer~\cite{2017arXiv170603762V}\footnote{The model default parameters are RNN hiddensize of $512$, CNN filtersize of $2048$, no of attention heads $8$.} has however eschewed recurrence in favour of attention mechanisms and established itself as the state of the art in neural sequence transduction based tasks. However, when the input sequence is corrupted or noisy, the underlying transductive model is not resilient to such noise. In Table~\ref{tab:motivation}, pre-processed ground truth information, corresponding noisy user inputs along with the model predictions from the Transformer model,  recurrent neural network (RNN) language model combined with an integrated spell-checker are shown in comparison to our model's predictions. We observe that the character level sequence-to-sequence transformer model fails to correct the noisy user input as well as the completion context. Word based RNN language model with integrated spell checker predicts better completion context if the input is not noisy and fails completely for noisy input. In contrast, our the predictions from our model that we discuss in the rest of the paper, has a low error rate in comparison with the ground truth.
\begin{table*}
	\centering
	\begin{tabular}{l c c c }
		\\
		\cmidrule(r){1-4}
		{\textit{Model}}
		& { \textit{Ground Truth}}
		& { \textit{User Input}}
		& {\textit{Model Prediction}}\\
		\midrule
		\multirow{8}{*}{Transformer~\cite{2017arXiv170603762V}} 
		& could you try ringing her & couks you tru ringing her & coucks your ringhing ing  \\
		&  is that ok & is that ok & is that on\\
		&  thanks i will &  thanka i will & thank  i will\\
		& yes i am playing & yew i am playing & yew i amplaying \\
		& is not can i call you & if not can i call you & if not an cally\\
		&  no material impact &  no material impact & no material micat \\
		 \midrule
		 \multirow{8}{*}{LM+SpellCheck}
		 & could you try ringing her & couks you tru ringing her &  coke you ' ringing her \\
		 &  is that ok & is that ok & is that ok\\
		 &  thanks i will &  thanka i will & hanka i will\\
		 & yes i am playing & yew i am playing & yes i am playing \\
		 & is not can i call you & if not can i call you & if not can i call you\\
		 &  no material impact &  no material impact & no material impact\\
		 	\midrule
		 	\multirow{8}{*}{CCEAD (ours)}
		 	& could you try ringing her & couks you tru ringing her &  could you try ringing her \\
		 	&  is that ok & is that ok & is that ok\\
		 	&  thanks i will &  thanka i will & thanks  i will\\
		 	& yes i am playing & yew i am playing & yew i am playing \\
		 	& is not can i call you & if not can i call you & if not can i call you\\
		 	&  no material impact &  no material impact & no material impact \\
		\end{tabular}
	\caption{Qualitative comparison of predictions (top to bottom) 1. Non recurrent sequence-to-sequence attention model prediction~\cite{2017arXiv170603762V}  2. Sequence based recurrent language model with integrated spell check 3. Sequence-to-sequence convolutional gated recurrent encoder gated decoder CCEAD (Our model) }~\label{tab:motivation}
\end{table*}
The recurrent neural network character and word language models~\cite{kim2015character} are extremely good at finding long range dependencies and context in the data, but struggle when the data is noisy. This is due to the sparsity of the error patterns in the training data; not all the error patterns are repeated with enough frequency. On typical mobile touch-screen virtual keyboards, the average user speed of typing is $25$ words per minute with a character error rate of not more than $10\%$ on average~\cite{Vertanen}. With smaller touch-screen size and index of difficulty, the error rate increases, but not significantly. Synthetically injecting noise in the data does not model the user's typing in a realistic way and thus not a viable option for generalization purposes.

Furthermore, most traditional and neural natural language understanding and speech systems train on as many as billion words corpus over multiple days. This is to ensure sufficient coverage over large corpus ~\cite{Vertanen, mikolov2013distributed, chelba2013one} that needs to be trained on multiple GPUs with generous memory bandwidth~\cite{kim2015character}. Further, the memory footprint of the trained models is sufficiently high especially for inference and prediction under resource constrained embedded devices such as mobile phones, smart-watches and other wearable devices. This also does not facilitate on-device training and other user privacy related issues. 

Although, RNNs are efficient on sequential data, they are unable to detect subword information or morphemes: \emph{invested}, \emph{investing}, \emph{investment}, \emph{investments} should have structurally related embeddings in the vector space~\cite{mikolov2011empirical}. The similar reasoning applies to error patterns of deletion, swapping and insertions for example ~\emph{invstment}, \emph{incest}, \emph{investinng} that have small edit distances to the revised targets. Convolutational neural networks (CNN)s are capable of representing such hierarchical complex hidden overlapping patterns more efficiently than RNNs and are natural candidates for this type of problem. 
\subsection{Sequence-to-Sequence Model}
Sequence-to-sequence models have shown tremendous potential in enhancing vanilla recurrent networks for input-output sequence mapping tasks such as machine translation~\cite{sutskever2014sequence}. An input side encoder captures the representations in the data, while the decoder gets the representation from the encoder along with the input and outputs a corresponding mapping to the target language. Intuitively, this architectural set-up seems to naturally fit the regime of mapping noisy input to de-noised output, where the revision prediction can be treated as a different language and the task can be cast as Machine Translation. However, one has to point out the limitation of such sequence-to-sequence mapping architecture where the input and the output sizes are fixed beforehand. Unlike, vanilla recurrent language models, the basic sequence-to-sequence model cannot generate (sample) new text or predict the next word. This is an important feature for keyboard decoder to accurately predict intended completion.

\begin{figure*}
	\centering
	\includegraphics[width=1.0\textwidth]{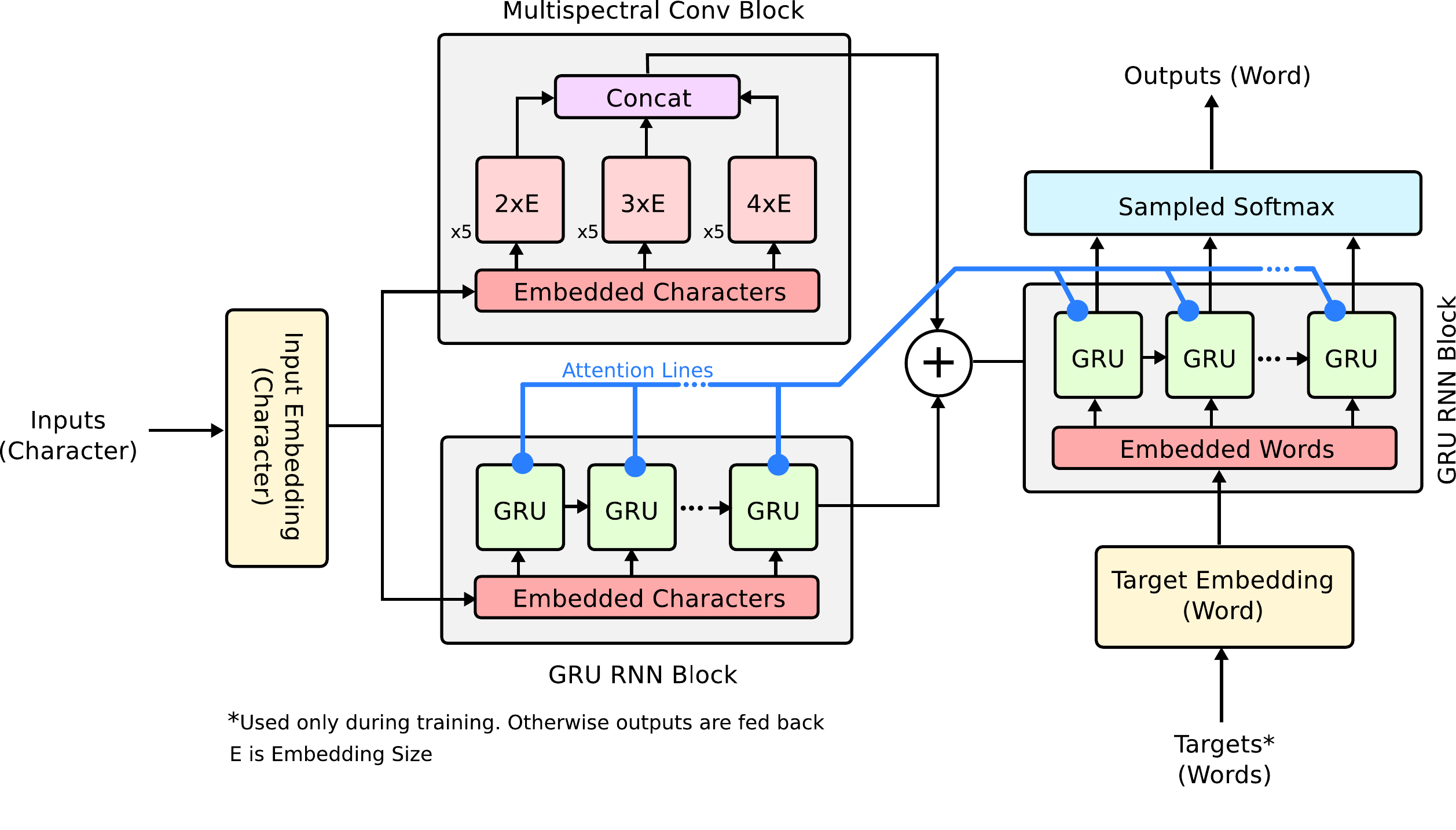}
	\caption{Architectural diagram of our character based convolutional gated recurrent encoder with word based gated recurrent decoder with attention (CCED). }~\label{fig:largeencdec}
\end{figure*}

Our approach in this paper is to address the correction and completion problem by having a separate corrector network as an encoder and an implicit language model as a decoder in a sequence-to-sequence architecture that trains end-to-end.
We propose a sequence-to-sequence text Correction and Completion Encoder Decoder Attention network (CCEAD) as illustrated in the architectural diagram in Figure~\ref{fig:largeencdec}. We use the smaller spoken English language Twitter Typo dataset of corrected tweets~\cite{twitter-typo}\footnote{http://luululu.com/tweet/} to induce noise in the larger conversational Open SubTitles dataset 2009~\cite{Tiedemann:RANLP5} of $50000$ unique words, by modelling the noise distribution over the Twitter data. Our proposed model trains on this combined dataset to capture the noisy to correct patterns whilst simultaneously learning the implicit language model. Our main contributions in this paper are summarized as follows:
\begin{itemize}
	\item
	We propose a character based sequence-to-sequence architecture with a convolutional neural network(CNN)-gated recurrent unit (GRU) encoder that captures error representations in noisy text.
	\item
	The decoder of our model is a word based gated recurrent unit (GRU) that gets its initial state from the character encoder and implicitly behaves like a language model.
	\item
	Our model learns conversational text correction and completions for use in any such text based system and is agnostic to the input medium.
	\item
	We perform experiments on synthetic, real datasets and real user noisy input data corresponding to stimuli sampled from the Enron email test datasets~\cite{vertanen2011versatile}.
	\item
	 We achieve an average character error rate (CER) of $2.6$ on real user data, that is competitive with the state-of-the-art keyboard decoder~\cite{Vertanen} with average CER of $1.6$ but with our model being approximately an order of magnitude smaller in memory footprint, training time and corpus size.
		\item
		We model the noise distribution over the twitter data~\cite{twitter-typo} and induce the noise in the OpenSubTitles 2009~\cite{Tiedemann:RANLP5} dataset; this combined dataset is used for training our model. 
		\item  We have released this novel dataset along with our trained model for inference and prediction which is available at the following url. We also plan to release the code associated with our model and all the other baselines as an open-source toolkit. 
		\item
		We report the baseline performance of our neural keyboard text decoder model over variety of alternative neural models.
\end{itemize}
\section{Related Work}
In the literature, likelihood probabilistic models have been used for text decoding, by the $n$-th order Markov assumption, the $n$-gram probabilities are estimated via counting and subsequent smoothing~\cite{chen1996empirical}. As count based models, the problem is in computing the probabilities
of rare word $n$-grams that can be incorrectly estimated due to data sparsity even with smoothing techniques. Also, the corpus size and training time is exponential in the size of the $n$-grams. Even the most sophisticated systems, the word level $n$ gram cannot exceed $5$ words due to the sheer size and complexity of processing. With the introduction of deep learning, the neural network architecture called RNNs has become popular for representing long term sequential data~\cite{greff2016lstm, sutskever2014sequence, cho2014learning, rodriguez1999recurrent}.
\begin{table}[t]
	\caption{Context based correction}
	\label{tab:context}
	\centering
	\begin{tabular}{ll}
		\cmidrule{1-2}
		Erroneous &   Truth  \\
		\midrule
		Say yello to him   & Say hello to him        \\
		Don't yello at me     & Don't yell at me       \\
		I prefer yello     & I prefer yellow         \\
		\bottomrule
	\end{tabular}
\end{table}
These networks were capable of modelling character based language models with much higher accuracy than the classical language models, and minimum assumptions on the sequence~\cite{mikolov2011empirical}. Further, the deep learning networks address the sparsity problem by parameterizing the words as vectors, such that the word embeddings obtained after training are semantically close words in the induced vector space of embeddings~\cite{kim2015character, bengio2003neural}. Convolutional Neural Networks (CNN)s~\cite{lecunhandwritten} are the feedforward networks that have achieved state-of-the-art results in computer vision. CNNs have also been applied to the natural language domain~\cite{kim2015character}; our model is different in the sense that we aim to correct and complete the text. Sequence to sequence learning~\cite{sutskever2014sequence} is used efficiently in the  machine translation tasks between languages, that developed over the previous architecture of encoder-decoder~\cite{cho2014learning}.
Despite the remarkable progress in deep learning, straight-forward tasks such as real-time word prediction, sentence completion and error correction, from text based decoding  perspective have received less traction in the research literature. One possible explanation for the less applicability of the deep learning solutions to these tasks is attributed to the resource constrained environments in which these embedded keyboard decoders operate, that also require low latency. The deep learning models have a large memory footprint and processing bandwidth requirement, where training can continue over days. Some commercially available touch-screen input systems such as SwiftKey uses deep learning. SwiftKey, Fleksy, and the built-in Android and iOS keyboards, address automatic error correction on a letter-by-letter and word-basis~\cite{Vertanen}; without much presence in the literature or open source community. This is one gap that we attempt to address in this work, we plan to release our code, training dataset and model performance as baselines for free access and furthering research.

Some of the classical techniques that dealt with error correction in the literature include combination of probabilistic touch model with character language model~\cite{goodman2002language}. A word level dictionary mapping based on geometric pattern matching~\cite{kristensson2005relaxing}, activity based error correction ~\cite{goel2012walktype} and sentence level revisions with fast entry rates~\cite{Vertanen}. Some models tackle both sentence completion and sentence correction~\cite{bi2014both}. 
Our work in this paper address both completion and correction; correction is performed at the character level complexity and completion is at the word level that leverage the rich encoder character representation. 
In the deep learning literature, although language models have been studied extensively in the form of character based language model, word based language model or character and word based language models~\cite{kim2015character}, these models cannot address the problem of simultaneous completion and correction due to the lack of available data; in particular dataset with enough noisy to correct mappings. Some work on recurrent deep learning networks have been applied to keyboard gesture decoding~\cite{alsharif2015long} over in-house collected data on Google keyboard that is not available openly.

There has been significant body of work in Natural Language Correction. 
To the best of our knowledge, this is the first research body of work on  deep neural networks for correction and completion. Our work uses sequence-to-sequence character embedding in the encoder with word based attention decoder, with openly accessible training data and the trained model for inference. We plan to release the code as open-source toolkit. In the work of Hasan et. al.~\cite{hasan2015spelling}, they perform statistical machine translation on character bi-grams on internal search query dataset. In the follow-up paper as a sequence-to-sequence model for the same task, their model trains on internal search queries as well, but works at both character and word level encoder-decoder architecture similar to ours. The difference is in the encoder architecture, where they have not used CNNs and RNN combined. Also, we do not use the more complex two layered and bi-directional LSTM RNN, instead we use single layer GRU RNN.

Although the work of Ziang et al. \cite{xie2016neural}, has similar character representations; both the encoder and the decoder operate at the character level. Further, the task is different from that of a traditional keyboard decoding spelling correction and completion, and can only operate on complete sentences.
In the follow-up work, Jianshu et al.~\cite{ji2017nested} uses nested attention for Grammatical Error Correction (GEC) by word and character level representation using sentence correction pairs. Their work is the most similar to ours in performing end-to-end learning on sentence correction pairs. However, the problem that we attempt to solve handles simultaneous correction and completion using an implicit neural language model that trains end-to-end in contrast to their model where a separate n-gram language model is integrated explicitly. Further, their model employs a word level sequence-to-sequence with attention and invokes the character level sequence-to-sequence nested attention only for OOV words. We handle OOV words as the encoder in our model is character based at all times.

The work on Natural Language Identification(NLI) deals with identification of the native language through means of spelling error correction~\cite{chen2017improving} which is a different task than ours and use character $n$-gram models. In the work of Bollman et al.~\cite{bollmann2017learning}, sequence-to sequence learning for German text normalization task is performed. The work by Schmaltz et al~\cite{schmaltz2017adapting}, performs grammatical corrections in sentences using a complete character based sequence to sequence model.

\begin{figure}
	\centering
	\includegraphics[width=1.1\columnwidth]{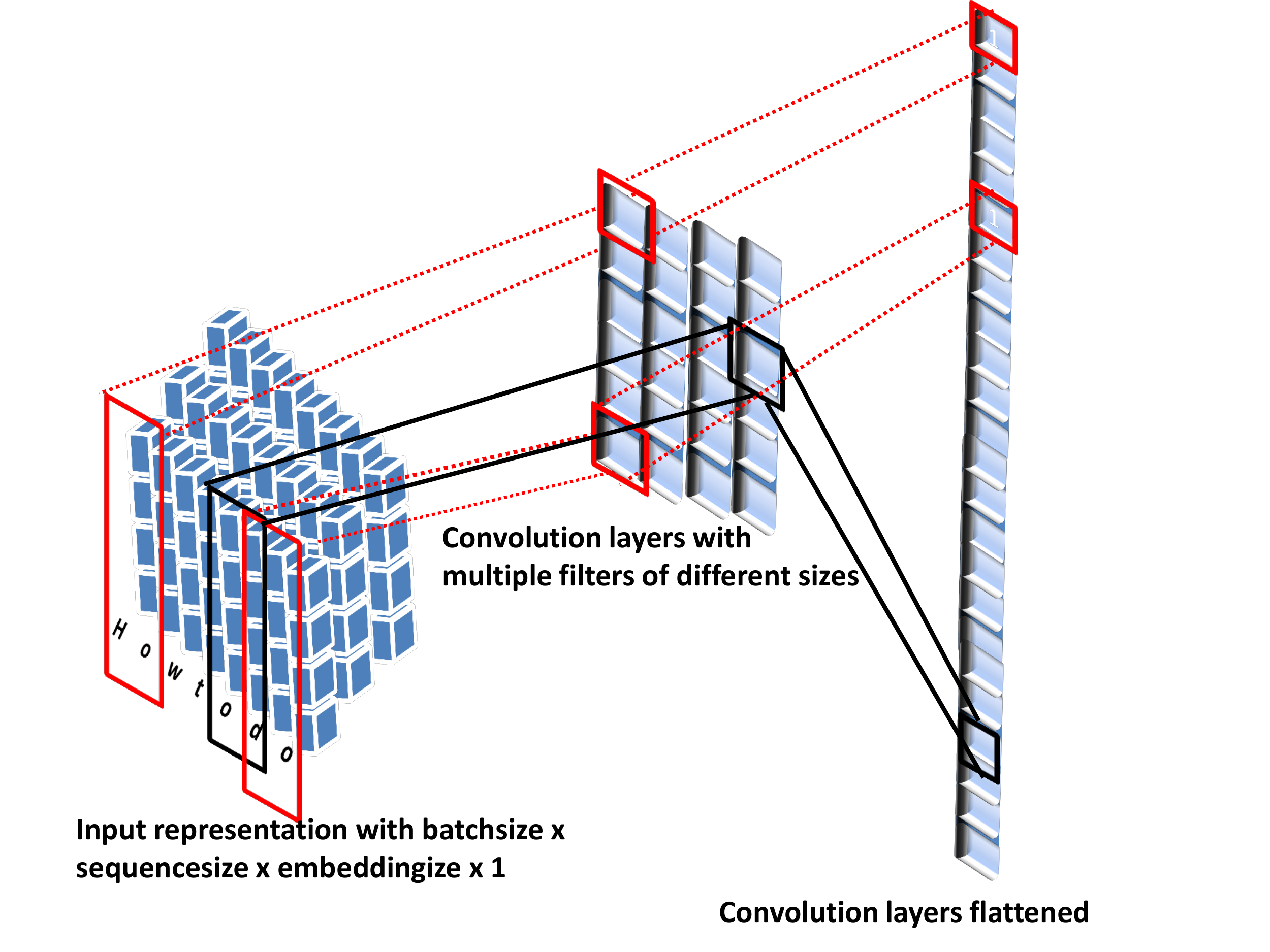}
	\caption{Illustration of the CNN module comprising the encoder of our CCEAD model used for capturing hidden representations in data }~\label{fig:cnn}
\end{figure}
\subsection{Notation}
Throughout the paper, we use $\textbf{W}\textbf{x}$ to denote matrix multiplication and $u \odot s$ denotes vector element wise multiplication or Hadamard product.
\section{Model Description}
The overall architecture of our model is illustrated in Figure~\ref{fig:largeencdec}. Our model has the similar underlying architecture of the sequence-to-sequence models used in machine translation. Sequence learning problems are challenging as deep neural networks (DNN) require the dimensionality of the inputs and targets to be fixed. Further, problems arise if the inputs are character concatenated representations whereas the outputs are word level representations. Therefore, in our model the encoder is composed of both recurrent and feed-forward units and operates at a character level whereas the decoder operates at a word level. Our model is composed on two main modules: Error Corrector Encoding Layer and Language Decoding Layer.
\subsection{Character Error Correcting Encoding Layer}
\subsubsection{Character Error Context Understanding}
Recurrent Neural Networks~\cite{werbos1990backpropagation, rumelhart1988learning} are extremely efficient in capturing contextual patterns.
For a sequence of inputs $(x_1, . . . , x_T )$, a classical RNN computes a sequence of outputs $(y_1, . . . , y_T )$ iteratively using the following equation:
\begin{align*}
\textbf{h}_t &= \sigma (\textbf{W}^{hx}\textbf{x}_t + \textbf{W}^{hh}\textbf{h}_{t-1} + \textbf{b}^{h})\\
\textbf{y}_t &= \textbf{W}^{yh}\textbf{h}_t.
\end{align*}
The power of RNNs lie in the ease with which it can map input sequence to target sequence, when the alignment between sequence is known ahead of time. In case of unaligned sequences, to circumvent the problem, the input and target sequences are padded to fixed length vectors, then one RNN is used as the encoder to map the padded input vector to a different fixed sized target vector using another RNN~\cite{cho2014learning}.
\begin{figure}[t!]
	\centering
	\subfloat {\includegraphics[scale = 0.45,trim=270 30 270 70]{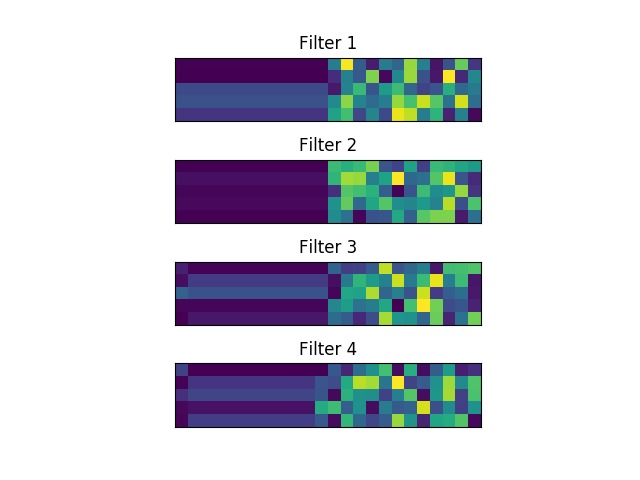}}
	\caption{CNN output from our CCEAD model visualized for activations as obtained from all four filters for the input {\tt how are you}. The activations are shown in light colours. From left to right, the input is represented as padding characters followed by {\tt  <EOS> <PAD> <PAD> \dots <PAD> u o y e r a w o h}. Best viewed in colour.} \label{cnn1visualizing}
\end{figure}
However, RNNs struggle to cope with long term dependency in the data due to vanishing gradient problem~\cite{bengio2003neural}. This problem is solved using Long Short Term Memory (LSTM) recurrent neural networks. However, for purposes of error correction, medium to short term dependencies are more useful. Therefore, our candidate for  contextual error correction encoding layer is the Gated Recurrent Network (GRU) which has similar performance to that of LSTM.
We evaluated a vanilla sequence to sequence model that we trained on exhaustive set of synthetic noisy to true mappings, without any context. The models could not generalize to all types of errors especially insertion and deletion.  
The example in Table~\ref{tab:context} highlights the importance of context for each of the sentences, that will require different corrections.
\begin{figure*}
	\centering
	\subfloat[ {\tt how are yu}]
	{\includegraphics[scale=0.33, trim=240 30 120 80]{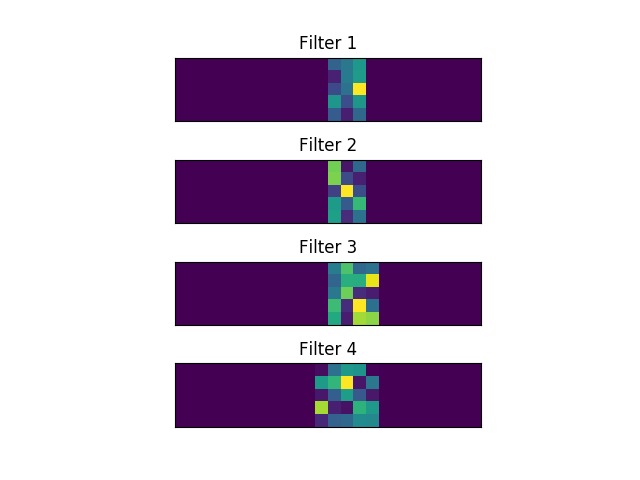}}	
	\subfloat[{\tt how are ypu}]
	{\includegraphics[scale=0.33, trim=40 30 120 80]{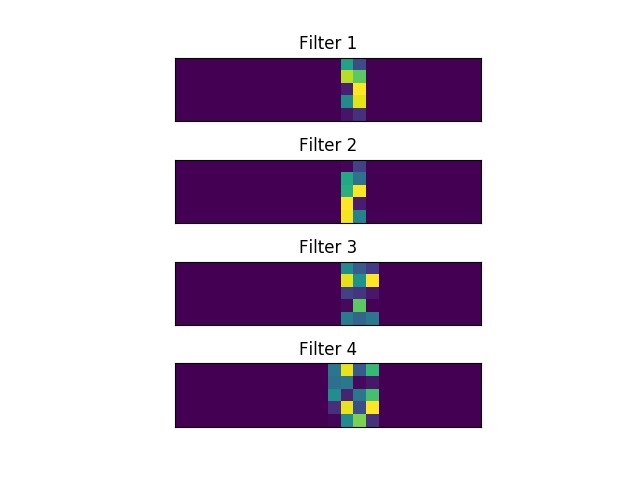}}
	\subfloat[{\tt how are yiu}]
	{\includegraphics[ scale=0.33, trim=40 30 120 80]{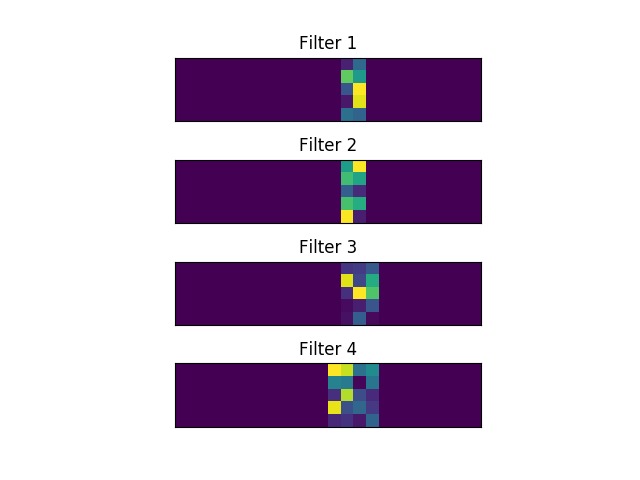}}
	
	\caption{Convolutional Neural Networks (CNN)'s output as part of the encoder state as visualized over three samples. Left: Convolution over {\tt how are yu}, centre: Convolution over {\tt how are ypu}, right: Convolution over {\tt how are yiu}. Best viewed in colour. Activations are indicated in lighter colours; the higher the activation, the lighter the colour. Convolution filter of width $k=2$, activates region of width $4$ for ambiguous error {\tt yu} to gather more context information over non-ambiguous errors where it activates smaller region. The visualizations are obtained from the trained model with response to a stimulus sample. }\label{fig:cnnvis2}
\end{figure*}
Gated Recurrent Unit (GRU)~\cite{cho2014learning, bahdanau2014neural, hochreiter1997long} are the fundamental units of our model. In GRU, the input and the hidden states are of the same size, that enables better generalization. The main motivation in using GRU over Long Short Term Memory networks (LSTM) as our fundamental unit is the size equality between input and state which is not the case with LSTM. Also, GRU and LSTM have exhibited similar performance across many tasks~\cite{chung2014empirical}. If the input vector is $x$ representing a character, and the current state vector is $s$, then the output is given as in Kaiser et. al.~\cite{kaiser2015neural}:
\begin{align*}
\textbf{h}_t &= \textbf{u}_t \odot \textbf{h}_{t-1} \\&+ \left({\textbf{1}-\textbf{u}_{t}}\right) \odot \tanh \left({\textbf{W}^{hx}\textbf{x}_{t} + \textbf{U}^{hh} \left({\textbf{r}_t \odot \textbf{h}_{t-1}}\right)} + \textbf{b}^{h}\right),
\end{align*}
where, 
\begin{equation*}
\textbf{u}_t = \sigma\left({\textbf{W}^{ux}\textbf{x}_t + \textbf{U}^{uu}\textbf{h}_{t-1} + \textbf{b}^{u}}\right) 
\end{equation*}
and,
\begin{equation*}
\textbf{r}_t = \sigma\left({	\textbf{W}^{rx}\textbf{x}_t + \textbf{U}^{rr}\textbf{h}_{t-1} + \textbf{b}^{r}}\right).
\end{equation*}
In the GRU equations above, $W^{*}$s and the $U^{*}$s are weight matrices with respect to the gates and hidden state while the $b^{*}$s are the bias vectors; where both of these type of  parameters are learnt by the model. $u$ and $r$ are the \emph{gates} as their elements have values between $\left[{0,1}\right]$ - $u$ is the update gate whereas $r$ is the reset gate~\cite{kaiser2015neural}. At every time step in the recurrent neural network, a GRU unit passes the result as the new state to the next GRU and to the output of the current time step.
\begin{figure*}[t!]
	\centering
	\subfloat[Embeddings region a.\label{fig:1a}] {\includegraphics[width=0.5\textwidth]{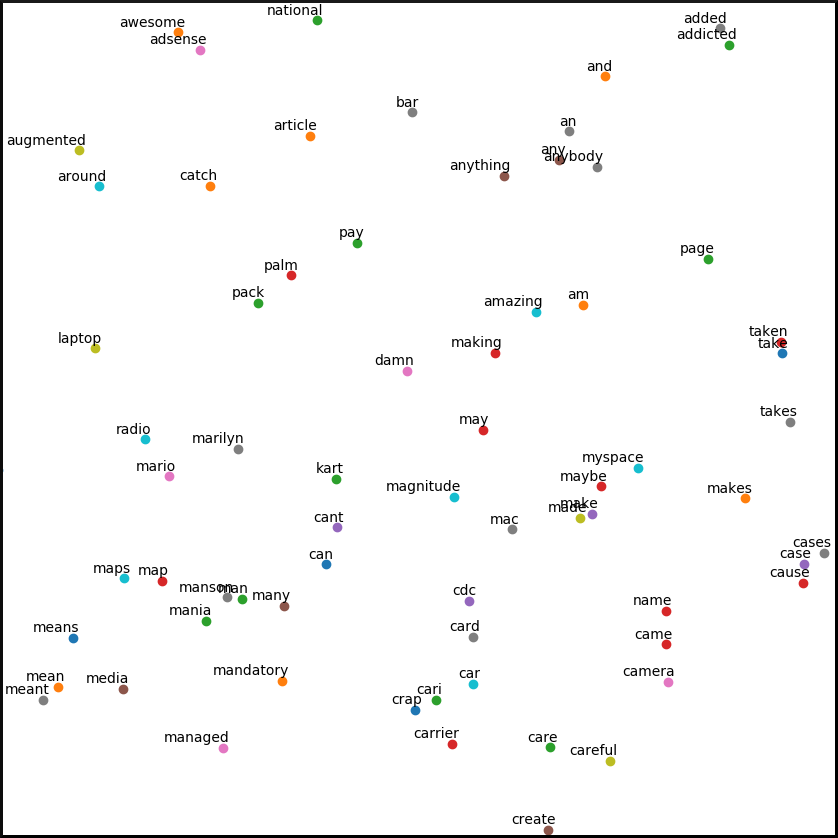}}
	\caption{Our model CCEAD's encoder state output embedding visualization afterthe  model has been trained using TSNE~\cite{maaten2008visualizing}. We observe similar words occupying closer regions in the induced embedding vector space.} \label{fig:encemb}
\end{figure*}
\subsubsection{Character Error Representation}
 The CNN encoding layer is shown in Figure~\ref{fig:cnn}. The convolutional layers are good at capturing the representations for insertion and deletion spelling errors discussed in previous sections. 
Both the GRU and the CNN are character based. For a sequence of, the input to the GRU and the CNN is simultaneous and in the form of a concatenated and padded fixed sized vector corresponding to the sequence length. CNN architecture applied to natural language processing  typically model temporal rather than spatial convolutions.

The characters considered in our model consists of $68$ characters, including $26$ English letters, padding symbol {\tt <PAD>} for aligning the sequences across batches of data input to our model, beginning of sequence marker {\tt <GO>} and end of sequence marker {\tt <EOS>} among other special characters as listed below:

\begin{spverbatim}
{0: '\t', 1: '\n', 2: '\r', 3: ' ', 4: '!', 5: '"', 6: '#', 7: '\$', 8: '\%', 9: '\&', 10: "'", 11: '(', 12: ')', 13: '*', 14: '+', 15: ',', 16: '.', 17: '/', 18: '0', 19: '1', 20: '2', 21: '3', 22: '4', 23: '5', 24: '6', 25: '7', 26: '8', 27: '9', 28: ':', 29: ';', 30: '=', 31: '>', 32: '?', 33: '@', 34: '[', 35: ']', 36: '_', 37: '`', 38: 'a', 39: 'b', 40: 'c', 41: 'd', 42: 'e', 43: 'f', 44: 'g', 45: 'h', 46: 'i', 47: 'j', 48: 'k', 49: 'l', 50: 'm', 51: 'n', 52: 'o', 53: 'p', 54: 'q', 55: 'r', 56: 's', 57: 't', 58: 'u', 59: 'v', 60: 'w', 61: 'x', 62: 'y', 63: 'z', 64: '{', 65: '|', 66: '}', 67: '<GO>', 68: '<PAD>'}
\end{spverbatim}

Suppose the length of the character sequence is given by $l$ for the $k$th word $w_k$, $\mathcal{V}_c$ is the vocabulary of characters and $\textbf{E}_c$ is the character embedding matrix of embedding dimension $d$ such that $\textbf{E}_c \in \mathbb{R}^{|{\mathcal{V}_c}|\times d}$. If word $w_k$ is made up of characters $[c_1, c_2, \dots, c_l]$, then the character level representation for word $w_k$ is $\textbf{V}_{c}^{k} \in \mathbb{R}^{l \times d}$. Central to temporal convolutions is 1D convolution. For input function $g(x) \in [1,l] \to \mathbb{R}$ and kernel function $f(x) \in [1,k] \to \mathbb{R}$, the convolution is  $h(y) \in [\ 1, \lfloor{(l-k+1)/d}\rfloor]\ \to \mathbb{R} $, for stride $d$ is defined as\cite{zhang2015character}:
\begin{equation*}
h(y) = \sum_{x=1}^{k}f(x)\cdot g(y.d - x + c),
\end{equation*}
where $c$ is the offset constant given by $c=k-d+1$, and $k$ is the width of the filter. There will be multiple filters of particular widths to produce the feature map, which is then concatenated and flattened for further processing.

Here, sequence length signifies how many characters define the context for the error for the GRU and is about $5$ characters or time steps. This sequence size will get adjusted to fixed sized length with padding to accommodate the shorter sequences. The number of neurons or cell size in the GRU is set to $256$. The CNN consists of $5$ filters with sizes varying in the range of $[2,3,4]$. The batch size indicating the number of instances to train on per batch is fixed to 100. The character vocabulary size is $30$ including the character used for padding, newline, space, and start of sequence.
The number of layers in the GRU is fixed to $1$.

\subsection{Word Level Decoder - Implicit Language Model}
The decoder is also a GRU recurrent network that does word based processing. The output from the encoder is a linear transformation between the final hidden state of the char based GRU in the encoder and the output of the fully connected layer of the CNN. This state is then used to initialize the state of the decoder. The decoder sees as input a padded fixed length vector of integer indexes corresponding to the word in the vocabulary. The input sequence is a sequence of words prefixed with the start token. The sequence length is set to $5$ which will then be padded for shorter sequences to generate fixed sized sequence. The vocabulary size for word based decoder is only about $3000$ words. The encoder-decoder is trained end-to-end having a combined char-word based representation on the encoder and decoder side respectively. The size of the GRU cell is $256$ with one layer. 

\subsubsection{Context based Attention}
The decoder constructs the context by attending to the encoder states according to attention mechanisms~\cite{bahdanau2014neural}. If the context is given by $\textbf{c}_i$, decoder state, previous encoder states by $\textbf{h}_i$ by $\textbf{s}_{i-1}$, then for the sequence $i, \dots, T$:
\begin{align*}
\textbf{c}_i &= \sum_{j=1}^{T}\boldsymbol{\alpha}_{ij}\textbf{h}_j\\
\boldsymbol{\alpha_{ij}} &= \frac{\exp(\textbf{d}_{ij})}{\sum_{k=1}^T\exp({\textbf{d}_{ik}})}\\
d_{ij} &= 	\text{tanh}(\textbf{W}_{s}\textbf{s}_{i-1} + \textbf{W}_{h}\textbf{h}_j + \textbf{b})
\end{align*}

Let $\mathcal{V}_w$ be the fixed size vocabulary of words. The decoder in our model behaves like an implicit language model by specifying a conditional distribution over $w_{t+1}$ consistent with the sequence seen so far $w_{1:t} = [w_1, \dots, w_t]$. The GRU recurrent unit achieves this by the application of the affine transformation of the hidden state followed by projection onto the word vocabulary by performing a softmax:
\begin{equation*}
Pr(w_{t+1}=j|w_{1:t}) = \frac{\exp(\textbf{h}_t\cdot \textbf{P}_j + b_j}{\sum_{i \in \mathcal{V}_w}\exp(\textbf{h}_t \cdot \textbf{P}_i + b_i)},
\end{equation*}
where $\textbf{v}_j$ is the $j$-th column of $\textbf{P}_{j}^{m \times \mathcal{|V|}_w}$, known as the output embedding and $b$ is the bias. Similar to our encoder side character embedding, our decoder takes word embeddings as inputs, for word $w_t$ at time $t$, $w_t=k$, the input to the decoder is the $k$ column of the word embedding matrix $\textbf{E}_w \in \mathbb{R}^{{\mathcal{|V|}_w}\times n}$.
During optimization the loss is evaluated on the output of the decoder at the word level. The loss function is the cross-entropy loss per time step summed over the output sequence $y$:
\begin{equation*}
L(x,y) = -\sum_{t=1}^{T}\log P(y_t|x, y_{<t}).
\end{equation*}

\section{Experiments}
\subsection{Dataset and Training Strategy}
\label{sec:data}
Due to the lack of representative data or data with sparse error patterns, we have build a dataset that our model is trained on. We use the Twitter typo dataset~\cite{twitter-typo} to induce the noise in the OpenSubTitles $2009$ movie dialog conversational dataset~\cite{Tiedemann:RANLP5}, for the common words between the two datasets. The Twitter typo corpus is a collection of spelling errors in tweets, where the data format is in pairs of a typo and its original form. The type of errors include insertion, deletion and substitution of one character. The Twitter typo corpus has $39,172$ spelling mistakes extracted from English Tweets along with their corresponding corrections. The manually corrected mistakes has an accompanying context word on both sides. The dataset was pre-processed to extract the original tweet itself with the accompanying revised tweet and a dictionary of typos to revised words created. 
The OpenSubtitles $2009$ data has subtitles from $30$ languages with a total number of files of $20,400$, total number of tokens $149.44M$ and total number of sentence fragments of $22.27M$. For every word, that is in the typo dictionary that we construct from Twitter, we replace the correct version of the word in the OpenSubtitles with the spelling error obtained from Twitter Typo.

The pre-processing stage cleans the OpenSubTitles dataset by running through a dictionary.
The resultant combined dataset which we call {\tt OpenTypo} comprises $50000$ words with every line containing a different sentence of approximately $7$ words and $27$ characters. The total number of sentences is approximately $2M$ with number of unique tokens being $12M$. Our model is trained on this combined dataset. The character set for the encoder has $68$ characters including the letters {\tt a-z}, numbers {0-9}, start of sequence symbol, end of sequence symbol, tab, newline, carriage return and single quotation among others. The decoder vocabulary is restricted to $50000$ words. The data is all converted to lower case.

Both the encoder and the decoder are jointly trained end to end on the {\tt OpenTypo dataset}. The output from the decoder is a softmax over the word vocabulary. The loss function used is the cross entropy loss for non-overlapping classes. We use Adams~\cite{Adam} optimization for training our model. The character and word embedding dimension are fixed to $200$. The training, test and validation data are split randomly. The accuracies from running the different baseline models in comparison to our model are reported in Table~\ref{tab:errortwitter}. 

\begin{figure*}[t!]
	\centering
	\subfloat[Replacement Edits \label{fig:errordist_twitter}] {\includegraphics[width=0.3\textwidth]{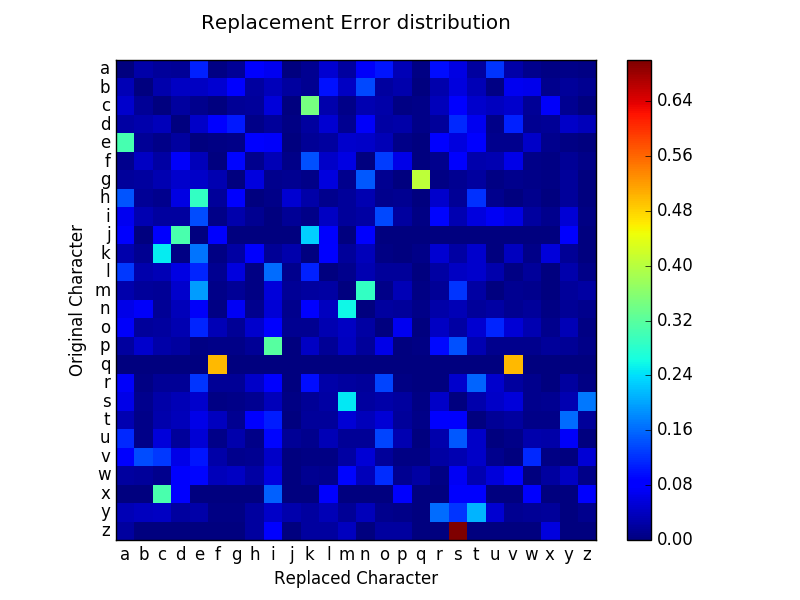}}
	\subfloat[Insertion Edits \label{errordist_twitter2}]
	{\includegraphics[width=0.3\textwidth]{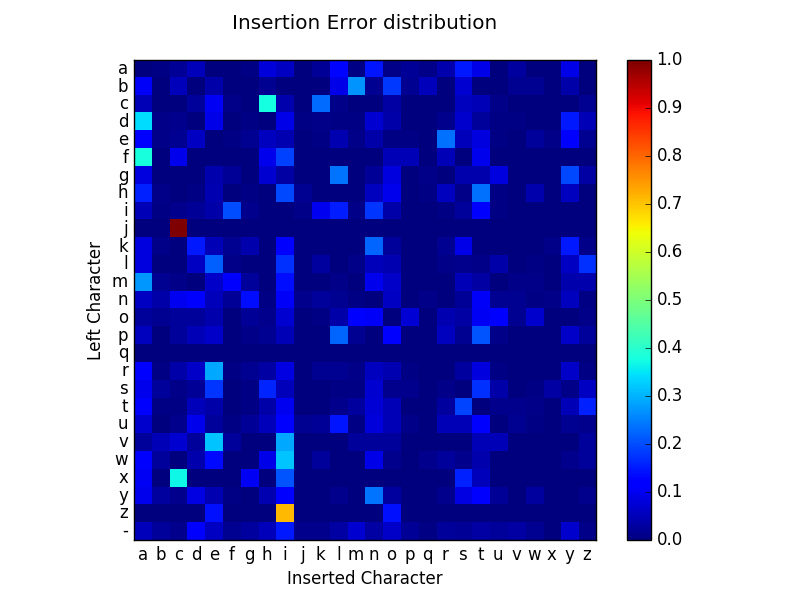}}
	\subfloat[Deletion Edits
	\label{errordist_twitter3}]
	{\includegraphics[width=0.3\textwidth]{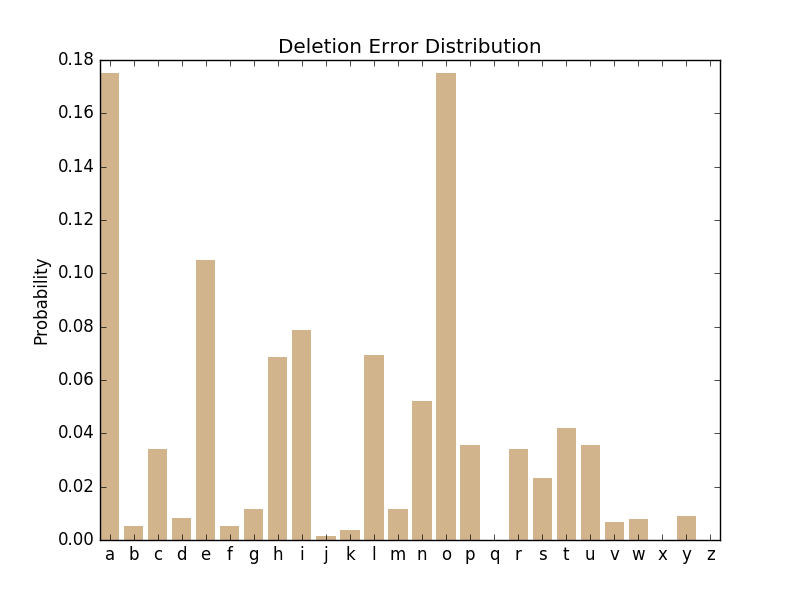}}
	\caption{Character Error Distribution on Twitter Typo Dataset. In (a) probability distribution of the original character being replaced by other characters (b) probability distribution of the characters to be inserted before left character (c) probability distribution of the character to be deleted.  } \label{fig:twiterror}
\end{figure*}

Figure~\ref{fig:twiterror} shows the noise distribution over the Twitter Typo dataset. The errors captured are namely over insertion of new character, substitution of existing character and deletion of existing character. 

{\em Synthetic Dataset}
We construct a synthetic dataset with manually induced noise by randomly sampling from a $2$D Gaussian (with standard deviation of $1$) centered on the key location of a simulated keyboard. We use the top $3000$ most common words in English \footnote{http://www.ef.com/english-resources/english-vocabulary/top-3000-words/}{\tt ef3000} and add noise for every letter constituting each word in the dictionary. The target for the noisy word thus created is the revised spelling of the word. The resultant dataset thus created has $79339$ noisy to true mappings. This dataset forms the input to our baseline models in Table~\ref{tab:error}. The experiments in this section evaluate the performance of {\tt CNN}, {\tt CNN-RNN-Parallel}, and {\tt CNN-GRU-Seq2Seq} models for correcting errors in noisy inputs without the presence of context in the data. The training set size size is $71405$ and the validation set has $3966$ instances.

\subsection{Baseline Models}
 Our initial experiments in Table~\ref{tab:error} over the synthetic dataset analyse the representative power of convolution neural networks (CNN). In all our results, the baselines and our model learn over the train partition, get tuned over the dev partition and are evaluated over the test partition for all the datasets respectively. All the code is written in using Tensorflow~\cite{abadi2016tensorflow} deep learning library version $1.2.1$. We perform experiments with different variations of convolutional filters in the number of filters, size of filters and strides. We also try {\tt CNN-RNN} and {\tt CNN-GRU} encoder decoder baselines. Both these networks have $256$ recurrent units. All the models compared in this set of experiments look at batch sizes of $50$ words and a sequence size of $5$ characters with a learning rate of $0.0002$ and run for $15$ epochs. The size of the network is detailed in Table~\ref{tab:error}. Throughout, when we refer to RNN in our baseline model instead of GRU we use two layered LSTM as the RNN.
 \begin{figure}
 	\centering
 	\includegraphics[width=0.4\textwidth]{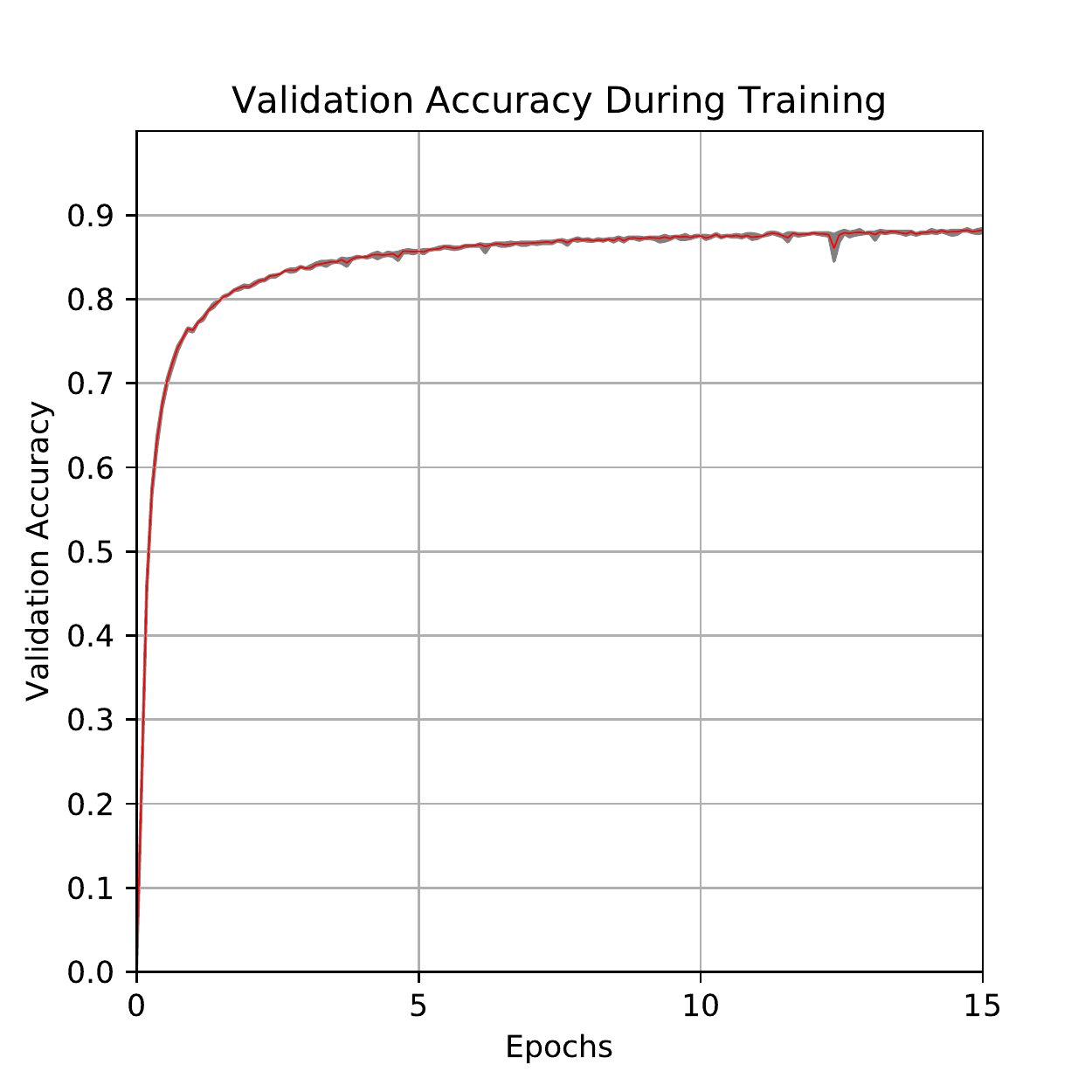}
 	\caption{Validation performance across iterations over Twitter Typo Dataset~\cite{twitter-typo}}~\label{fig:lossplot}
 \end{figure}
 In Table~\ref{tab:errortwitter}, our baselines are character based sequence to sequence models: {\tt CNN-RNN}, {\tt CNN-GRU}, {\tt RNN-RNN}. Since character level models cannot predict at a word level, to compare with our CCEAD model, we have additional baseline of {\tt RNN-C2W} where the encoder works at a character level and the decoder functions at a word level as indicated by letters $C$ and $W$. Additionally, we also compare with the latest state-of-the-art sequence to sequence model with multiplicative attention called {\tt Transformer}~\cite{2017arXiv170603762V}. The model depicted as CCED in the paper is indicated as {\tt CNN-LSTM-C2W} in the Table~\ref{tab:errortwitter}. This is the identical model as our CCEAD proposed in this paper minus the attention mechanism. The C2W models outputs by performing a softmax over the vocabulary. Our model has a learning rate of $0.002$, batch size of $100$, sequence length of $5$, embedding size of $200$, number of filters $5$, dropout rate of $0.3$ using Adams optimization.  
 The state-of-the-art Transformer model has a hidden size of $512$, batch size of $4096$, sequence length of $256$, number of layers $6$ and learning rate of $0.1$. The recurrent neural language model baseline is a GRU word model having a cell size of $256$ and the rest of the hyperparameters same as our model. The language model is combined with an edit distance spell checker for making final predictions.
 The results reported in Table~\ref{tab:error} and Table~\ref{tab:errortwitter} are the highest word level accuracy over all epochs on the test partition of the synthetic and the {\tt TwitterTypo} dataset respectively. In Table~\ref{tab:errorfused}, we report the results of our model's predictions over the {\tt OpenTypo} dataset. Training over {\tt OpenTypo} takes less than a day to complete, on a $12$GB NVIDIA TitanX GPU. The accuracy that we report is word accuracy and accuracy measured over the entire sequence. 
 \begin{table}
 	\centering
 	\begin{tabular}{l c c c c c}
 		& & \multicolumn{4}{c}{\small{\textbf{Model Hyperparamters}}} \\
 		\cmidrule(r){2-6}
 		{\textit{Model}}
 		& { \textit{Filters}}
 		& { \textit{Widths}}
 		& { \textit{Stride}}
 		& { \textit{RNN Size}}
 		& { \textit{Char Acc.}} \\
 		\midrule
 		CNN & 5 & [1-7] & 1 & - & 0.863\\
 		CNN-Stride& 3 & [2-5] & 2 & - &  0.483\\
 		CNN-Filters& 10 & [2-7] & 1 & - & 0.557\\
 		CNN-GRU & 3 & [2-5] & 1 & 256 & 0.986\\
 	\end{tabular}
 	\caption{Various sequence-to-sequence baselines' performance on our synthetic dataset for noisy to correct word mappings without any contextual information. All baselines are character based models. The accuracy is reported on test set is over the entire sequence}~\label{tab:error}
 \end{table}
 
 \begin{table}
 	\centering
 	\begin{tabular}{l c c c c}
 		& & \multicolumn{3}{c}{{\textbf{Model Hyperparameters}}} \\
 		\cmidrule(r){3-5}
 		{\textit{Model}}
 		& { \textit{Filters}}
 		& { \textit{Widths}}
 		& { \textit{RNN Size}}
 		& { \textit{Char/Word Acc.}}\\
 		\midrule
 		CNN-RNN & 5 & [2-10] &  256 & 0.91\\
 		CNN-GRU & 20 & [2-4] &  256 & 0.958\\
 		RNN-RNN &- & -  & 80 & 0.974\\
 	 		\midrule
 		RNN-C2W & -& - &  256 & 0.892\\
 	 	CNN-GRU-C2W & 5 & [2-5] & 256 & 0.904\\
 	 	
 	\end{tabular}
 	\caption{Sequence-to-sequence models evaluation on the TwitterTypo data with context. Baselines performance reported in the table above the horizontal line are character based while the ones below the line are character-to-word models including the CCED model CNN-LSTM-C2W . In the case of character based models, the accuracy is measured at a sequence level, whereas the word models report the word level accuracy.}~\label{tab:errortwitter}
 \end{table}
  \begin{table}
  	\centering
  	\resizebox{\columnwidth}{!}{%
  	\begin{tabular}{l c c c c c}
  		& & \multicolumn{4}{c}{{\textbf{Model Hyperparameters}}} \\
  		\cmidrule(r){2-6}
  		{\textit{Model}}
  		& { \textit{Filters}}
  		& { \textit{Widths}}
  		& { \textit{RNN Size}}
  		& { \textit{Word Acc.}}
  		&{ \textit{Seq Acc.}}\\
  		\midrule
  			Transformer~\cite{2017arXiv170603762V} &- & - & 512 & 0.787 & 0.558 \\
  				\midrule
  			LM+SC &- & - & 256 & 0.819 & \\
  				CNN-RNN-C2W & 5 & [2-5] & 128 & 0.968 & 0.616\\
  				CCEAD (Ours) & 5 & [2-5] & 128 & 0.981 & 0.689\\
  			\end{tabular}
  		}
  			\caption{Sequence-to-sequence model performance on the OpenTypo test parition with context (sentence length instances). Models depicted above the horizontal line in the Table are character based while the ones below the line are character-to-word models including the CCED model or CNN-LSTM-C2W as well as our proposed CCEAD model . In both cases, accuracy is measured over the entire input sequence. All models are trained on OpenTypo dataset.}~\label{tab:errorfused}
  		\end{table}	
\subsection{Analysis}
We visualize the word embeddings using TSNE~\cite{maaten2008visualizing} as learnt by our trained model, in Figure~\ref{fig:encemb}, for a subset of our vocabulary. We observe that the encoder state as visualized here has learnt to cluster similar words in the closely related regions of the vector embedding space for all embeddings. The embeddings are visualized using the output state of the encoder, for our model trained on Twitter Typo dataset. 
In Figure~\ref{cnn1visualizing}, the hidden representations from the encoder , specifically the CNN state is visualized with respect to a test sample using our trained model. The test sample when provided, passes through the CNN layer and just before the linear operation as shown in the architecture diagram in Figure~\ref{fig:largeencdec} indicated by a {\tt +} symbol, we capture the activations. In Figure~\ref{fig:cnnvis2}, the representations of the CNN filters corresponding to some noisy inputs are shown using our trained model. We see that when the noisy input is {\tt yu} with an edit distance of $2$, the activations are spread across a larger area even for the filter with size $2$ to likely gather more context. In contrast when the noisy input has an edit distance of $1$, the filter activations are concentrated on the letters that need attention.

In Figure~\ref{tab:error}, we report the results of the empirical evaluation of the comparison between different encoder baselines with our encoder model: {\tt CNN-GRU}, over the synthetic dataset that we constructed in Section~\ref{sec:data}.
 We observe that varying the convolutional layers by increasing the number of filters arbitrarily, increasing the stride, or the size of individual filters, causes the models to overfit.  In general, when the size of the individual filters vary between $[2-5]$, where most of the words in the dataset have length of $5$ characters, gives the highest accuracy. In Figure~\ref{tab:errortwitter}, we empirically evaluate the sequence-to-sequence baseline models with the sequence-to-sequence CCED model listed as {\tt CNN-LSTM-C2W} (our model without attention), over the Twitter Typo dataset. Although, the character based baselines report a higher character level accuracy over word models, they cannot be evaluated at a sentence level. Our model reports the highest word level accuracy in comparison with the baseline: {\tt RNN-C2W} without CNN in the encoder side. In Table~\ref{tab:errorfused}, we analyse the word level recurrent neural language model with spell check using edit distance and other state-of-the- art namely Transformer~\cite{2017arXiv170603762V}, in comparison with our model over the OpenTypo dataset. Our model CCEAD (with attention) outperforms all the character level state-of-the-art. This shows that edit distance based recurrent language models on its own cannot correct and complete efficiently.
 
\begin{figure}
	\centering
	\includegraphics[width=0.8\columnwidth]{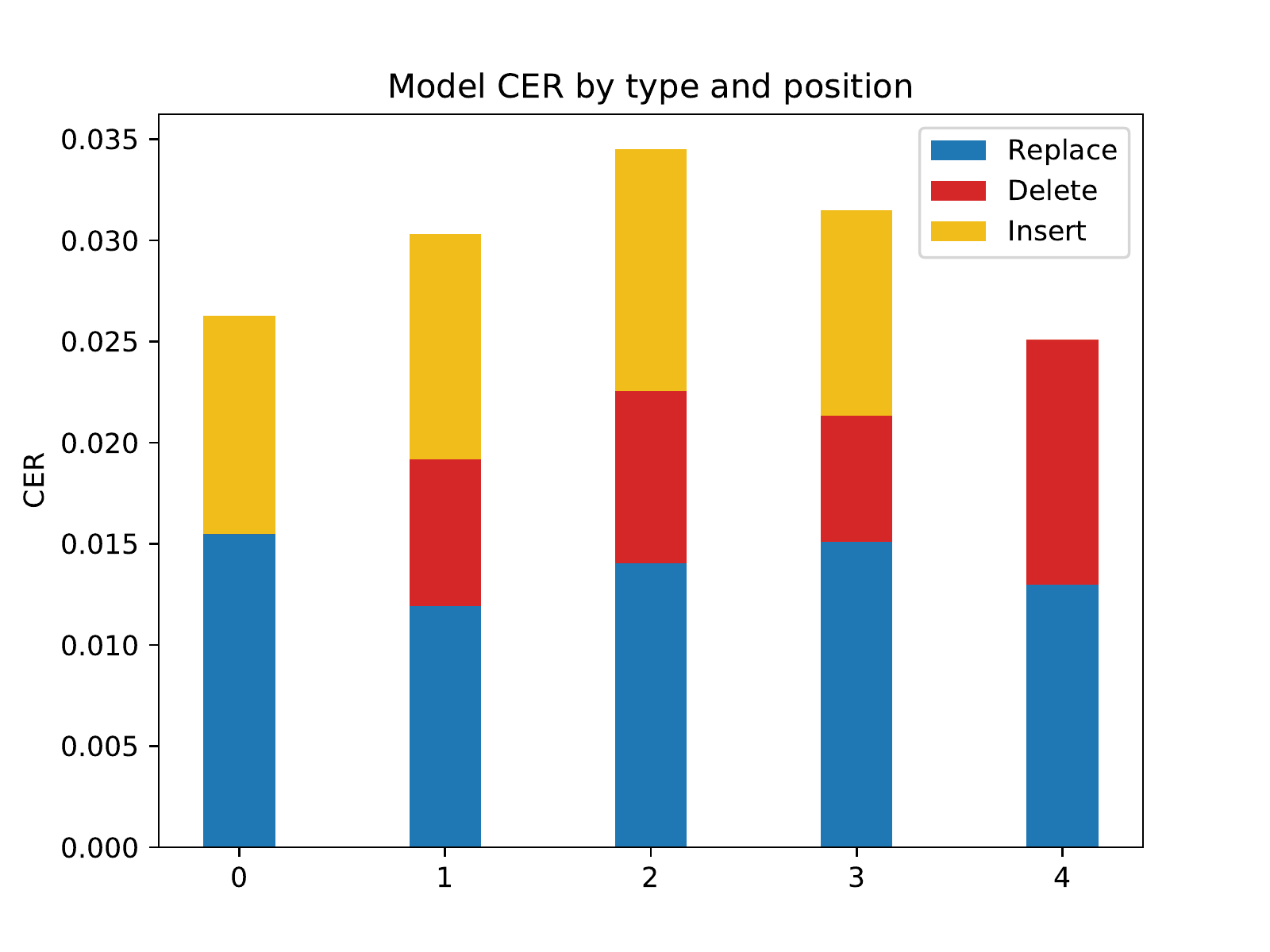}
	\caption{Character error rate for our model prediction by position on all five letter words over Twitter Typo test set}~\label{fig:cer_twitter}
\end{figure}

In Figure~\ref{fig:lossplot}, the validation accuracy of our CCEAD model is plotted over the training time required to converge on the {\tt OpenTypo dataset}. The accuracies are averaged over $5$ trials. In Figure~\ref{fig:cer_twitter}, we plot the character error rate of our model prediction across the TwitterTypo dataset for insertion, substitution and deletion errors with a maximum CER of $1.5\%$ for substitution errors on the first position of all five letter words.
\begin{table}
	\centering
	\begin{tabular}{l c c }
		\\
		\cmidrule(r){1-3}
		{\textit{Model}}
		& { \textit{Inputs}}
		& {\textit{Prediction}}\\
		\midrule
		\multirow{8}{*}{CNN-RNN} 
		& 'wavve' &'wave' \\
		& 'hpe' & 'hope' \\
		& 'are yiu' & 'are you' \\
		& 'thst os' & 'that is' \\
		& 'mpte yhe' & 'so much' \\
		& 'wulk sse' & 'you use' \\
		& 'will sse' & 'will use' \\
		& 'sge' & 'see'  \\
		
		\midrule
		\multirow{8}{*}{RNN-C2W} 
		& 'wavve' &'wave' \\
		& 'hpe' & 'hope' \\
		& 'are yiu' & 'are you' \\ 
		& 'thst os' & 'this is' \\ 
		& 'mpte yhe' & 'me the' \\
		& 'wulk sse' & 'would we' \\ 
		& 'will sse' & 'will see' \\ 
		& 'sge' & 'me' \\
		\midrule
		\multirow{8}{*}{CNN-GRU-C2W} 
		& 'wavve' &'wave' \\ 
		& 'hpe' & 'hope' \\  
		& 'are yiu' & 'are you' \\ 
		& 'thst os' & 'that is' \\ 
		& 'mpte yhe' & 'more the' \\  
		& 'wulk sse' & 'well be' \\ 
		& 'will sse' & 'will see'  \\ 
		& 'sge' & 'see'\\
		\midrule
	\end{tabular}
	\caption{Qualitative evaluation over test samples drawn from an arbitrary distribution.}~\label{tab:samples}
\end{table}
Table~\ref{tab:samples}, refers to the qualitative evaluation of some test samples drawn from an arbitrary distribution. Our closest model CCED (without attention) predicts with the best qualitative contextual correction in comparison to the other models. Here, CCED is trained on the Twitter Typo dataset only. Despite the superior performance of the character based baselines as shown in Figure~\ref{tab:errortwitter}, the qualitative evaluation shows that character based baselines do not product quality word level corrections for very noisy stimulus.
%
%
%
\subsection{User Study}
We performed user study with $8$ users. We developed an application for the Google Android Pixel mobile phone for using their touchscreen keyboard. We also developed a browser data input application for data entry using physical keyboard. The user is shown a stimulus text to type that is sampled at random from the Enron email dataset~\cite{vertanen2011versatile} and types the stimulus using the mobile touchscreen keyboard or the physical keyboard respectively without editing the typed text in any way. The results are reported in Table~\ref{tab:Userstudy} after performing an off-line analysis.

In the literature, keyboard decoder predictions are evaluated using the Levenstein distance or the edit distance. This measures the similarity between two strings; the source string (s) and the target string (t). The distance is the number of deletions, insertions, or substitutions required to transform $s$ into $t$. Using the original Levenstein distance, the edit distance between $s=s_1, \dots,s_n$ and $t=t_1, \dots, t_m$ is given by $d_{mn}$ calculated using the recurrence relation:
\begin{align}
\begin{split}
d_{i0} &= \sum_{k=1}^{i}w_{del}(t_k) \hspace{0.5cm} \text{for} \hspace{0.5cm} 1\leq i \leq m \\
d_{0j} &= \sum_{k=1}^{j}w_{ins}(s_k) \hspace{0.5cm} \text{for} \hspace{0.5cm} 1 \leq j \leq n \\
d_{ij} &= \begin{cases}
d_{i-1,j-1} \hspace{0.5cm}	\text{for} \hspace{0.5cm} s_j = t_i\\
\min\begin{cases}
d_{i-1,j} + w_{del}(y_i) \hspace{0.2cm} \text{for}\hspace{0.2cm}1 \leq i \leq m, 1 \leq j \leq n \\
d_{i,j-1} + w_{ins}(s_j)\hspace{0.5cm} \text{for}\hspace{0.2cm} s_j \neq t_i\\
d_{i-1,j-1} + w_{sub}(s_j,t_i)
\end{cases}
\end{cases}
\end{split}\label{eq:cer}
\end{align}

 Our model CCEAD (with attention) and the {\tt CNN-LSTM-C2W} CCED model (without attention) along with the baselines including the language model+spell check ({\tt LM+SC}) are trained on the {\tt OpenTypo} dataset. Also, the user input, predicted text and the ground truth are all normalized by removing capitalization.
For both mobile touch-screen and physical keyboard experiments, our test data is drawn from the mobile email dataset Enron~\cite{vertanen2011versatile}. We randomly sample from the $200$ sentences, for displaying the text to the user. The user types using the mobile touchscreen in-built keyboard (with predictions and auto-correct turned off) or the physical keyboard for browser based input method.

\begin{figure*}[t!]
	\centering
	\subfloat[Ideal words per minute] {\includegraphics[width=0.3\textwidth]{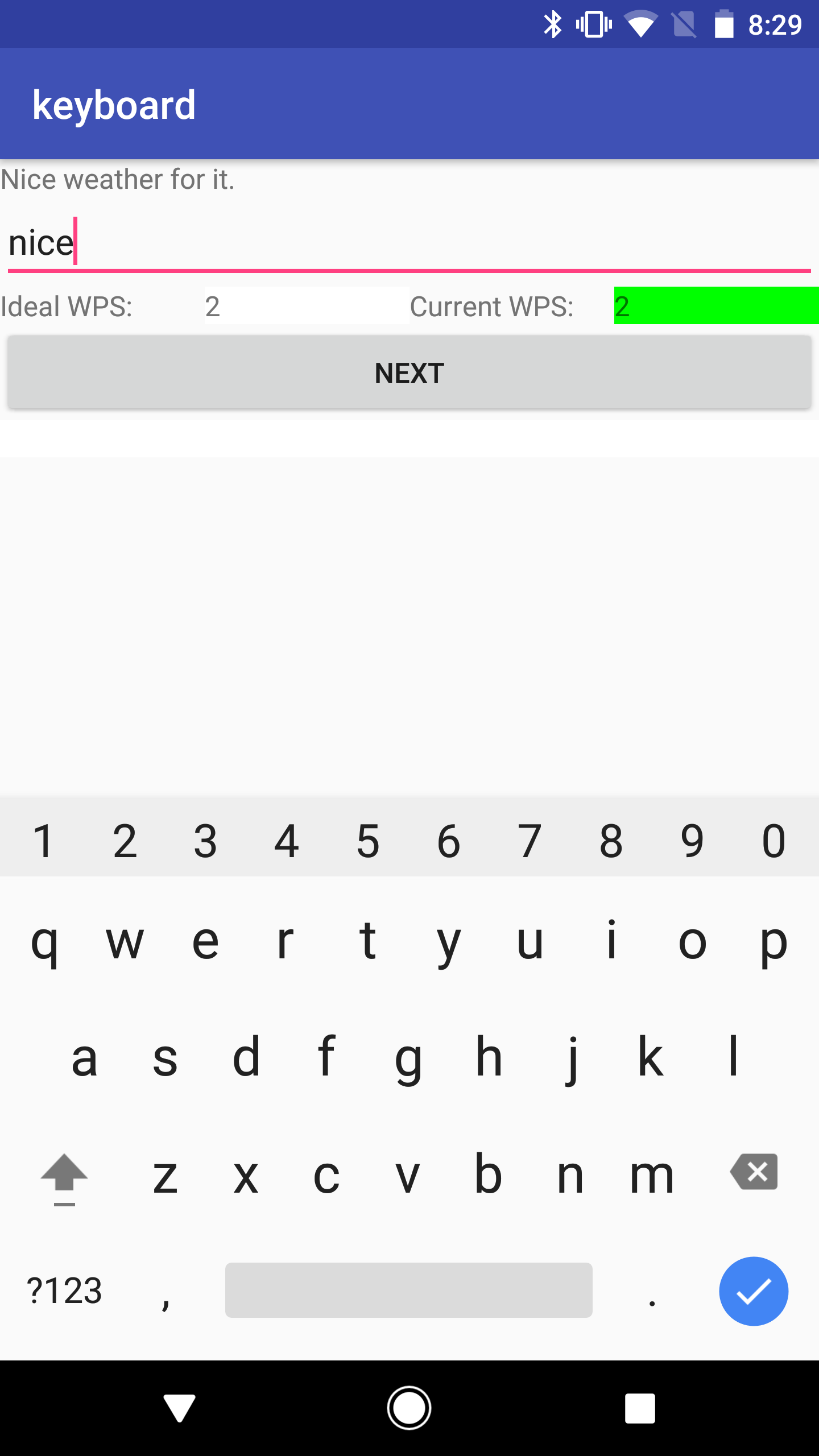}}\hspace{1cm}
	\subfloat[Non ideal words per minute]
	{\includegraphics[width=0.3\textwidth]{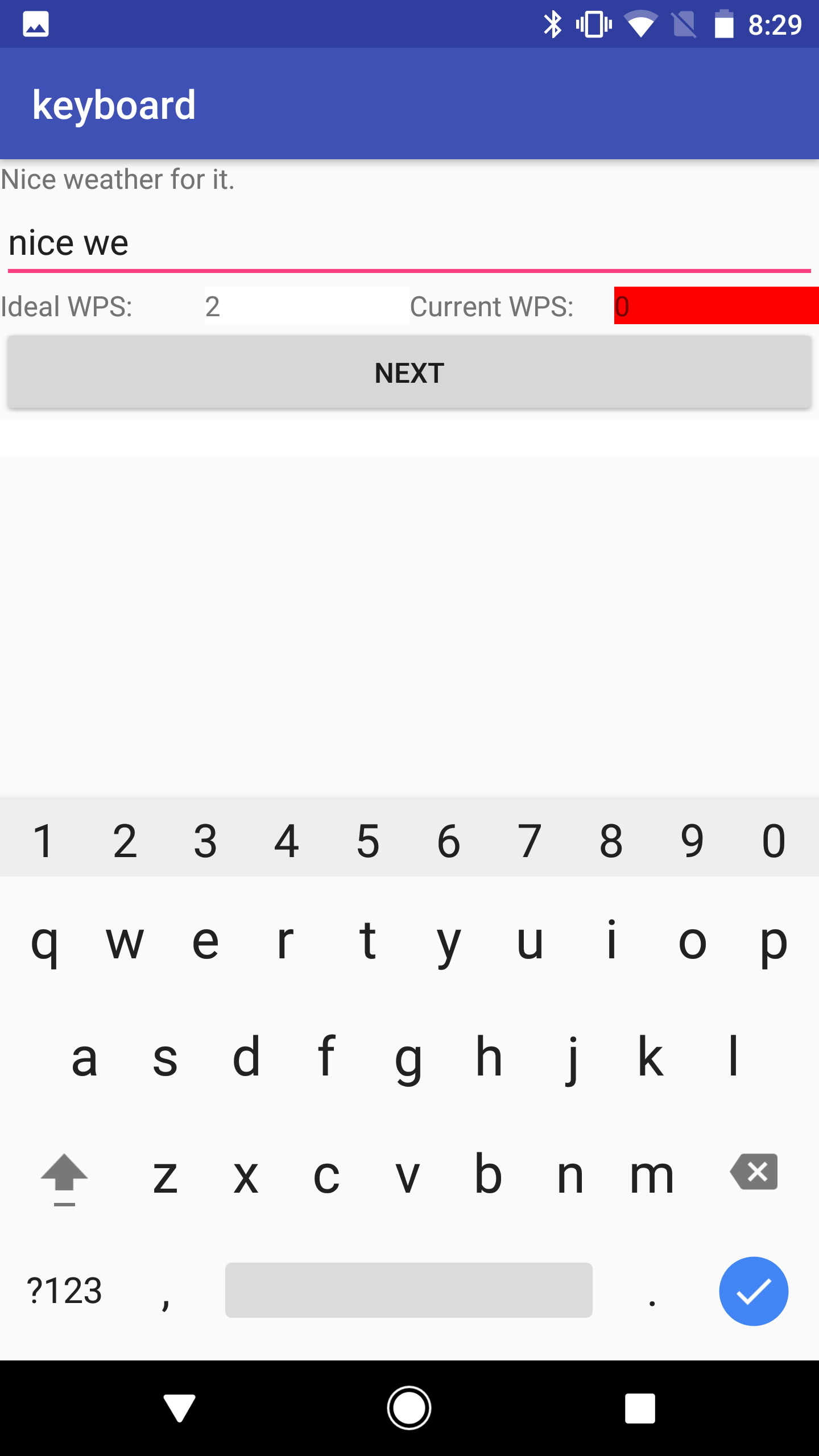}}
	\caption{Keyboard application for User typing data collection study on Google Pixel phone.  } \label{fig:android}
\end{figure*}

For the mobile touchscreen study, we have developed an Android application for Android $7.1.2$, for the Google Pixel $32$ GB phone. The application has a text box where the user can type. Above the text box, the text input to type is displayed. Although, the prompt text displays punctuations, the user does not have to enter punctuations. As soon as the user starts typing, an Ideal WPS (words per second) and Actual WPS values are displayed. This indicates the user's speed of typing. The user should not type very slowly, in which case the Actual WPS display turns red. Once the user finishes typing, the Next button is pressed, when the user's input is passed through the CER calculator for logging in the server. Only if the average CER is less than $10\%$, we accept the user input for logging purposes.
The Figure~\ref{fig:android} shows our Android application screen-shots while being used.
\begin{table*}[t]
	\caption{Offline decoding results on user inputs for test data~\cite{vertanen2011versatile} collected on Google Pixel mobile touchscreen keyboard and physical keyboard. The performance measure reported is {\tt CER} as given by Equation~\ref{eq:cer}. Here, we report the best performing result per user for all models. An empty cell means that particular model could not be evaluated using the corresponding input medium.}
	\label{tab:Userstudy}
	\centering
	\begin{tabular}{llllll}
		\cmidrule{1-6}
		User No.  & Input medium     &  CCED  & LM+SC  & CCEAD(Ours) & Velocitap~\cite{Vertanen}  \\
		\midrule
		User 1 & Mobile touch-screen & 1.4  & 1.0 & 1.0 & 1.0\\
		\midrule
		User 2 & Mobile touch-screen & 1.9  & 2.0 &  1.9& 1.9\\
		\midrule
		User 3 & Mobile touch-screen & 0.1  & 0.1 &  0.1 & 0.0\\
		\midrule
		User 4 & Mobile touch-screen & 0.8  & 1.3 &  1.0 & 1.0\\
	    \midrule
		User 5 & Mobile touch-screen & 1.1  & 1.3 &  0.8 &  1.0\\
		\midrule
		User 6 & Mobile touch-screen & 5.8  & 6.3  &  1.1 & 1.0\\
		\midrule
		User 7 & Mobile touch-screen & 0.8  & 0.9 &  0.8 & 1.0\\
		\midrule
		User 8 & Mobile touch-screen & 5.1  & 5.1 &  5.3 & 3.0 \\
		\midrule
		\midrule
		User 2 & Physical keyboard  & 1.1   & 1.4 &  1.3 & -\\
		\midrule
		User 3 & Physical keyboard  & 0.5   & 0.9 &  0.5& -\\
		\midrule
		User 4 & Physical keyboard  & 1.6   & 2.8 &  1.4 & - \\
		\midrule
		User 5 & Physical keyboard  & 1.6   & 1.9 &  1.6 & - \\
		\midrule
		User 6 & Physical keyboard & 3.9   & 3.8 & 4.1 & - \\
		\midrule
		User 7 & Physical keyboard & 3.3   & 3.3 &  3.2 & - \\
		\midrule
		User 8 & Physical keyboard & 4.9   &  5.1 &  4.9 & - \\
		\bottomrule
	\end{tabular}
\end{table*}

In Table~\ref{tab:Userstudy}, the user test data comparative evaluation is provided. The additional baselines used for comparative evaluation are described as follows. {\tt LM+CCED} is the CCED model of {\tt CNN-LSTM-C2W} (without attention) along with a separate word based language model that train end-to-end with the rest of the model. The predictions are a an affine transformation between the language model prediction and the {\tt C2W} prediction. {\tt Velocitap} is the state-of-the-art keyboard decoder~\cite{Vertanen}. Since {\tt Velocitap} functions as a keyboard model as well, here, we configure the Velocitap model to the Google Nexus keyboard layout. It is important to note that unlike the other baselines that we use here, Velocitap is not trained on the same dataset i.e. OpenTypo. We did not have access to train the Velocitap model on our dataset. According to the paper~\cite{Vertanen}, the model is trained on a $1$ billion word corpus with a language model of file size on disk of $1.5GB$. 

Table~\ref{tab:Userstudy} reports the results of the user study. Typing data is collected from users on a mobile touchscreen keyboard application and a physical keyboard browser application. Each user data is treated as a test set and the predictions of our proposed model, baselines and the state-of-the-art are evaluated. We observe that our model is consistent in performance over the baseline models. Our model is competitive with the state-of-the-art in terms of quality of predictions. However, in terms of the corpus size, memory footprint, parameters, training costs, we perform significantly better than the state-of-the-art as shown in Table~\ref{tab:comparemodels}. The failure mode (high CER) in the case of User $8$ for mobile touchscreen predictions is due to the presence of a lot a joined words in the input stimulus. Our noise model did not factor joining words, hence the data did not have joined words at all.

In Table~\ref{tab:Googlecomp}, we compare our model predictions with Google search predictions (using Google's own 'Search instead for'). It is very interesting to see our model's predictions supersede Google's predictions especially when our model is such a small scale and low resource implementation.

\begin{table*}[t]
	\centering
	\begin{tabular}{l c c c c}
		& & \multicolumn{3}{c}{{\textbf{Model Sizes}}} \\
		\cmidrule(r){3-5}
		{\textit{Model}}
		& { \textit{Corpus Size (words)}}
		& { \textit{No of Parameters}}
		& { \textit{Training Time (hours)}}
		& { \textit{Memory Footprint (Disk)}}\\
		\midrule
		 CCEAD (Ours)& 12M  & 17M
		  & 24   & 200MB\\
		 Velocitap~\cite{Vertanen}& 1.8B  &  - &  days &1.5GB \\
		 Transformer~\cite{2017arXiv170603762V}& 12M & 44M  & 2 & 541MB\\
		LM + SC & 12M & 26K  &  24 & 320MB\\

		CCED & 12M & 17M& 24 &200MB\\
	\end{tabular}
	\caption{Model comparison. Cell with a `-` is not applicable for the given model comparison.}~\label{tab:comparemodels}
\end{table*}

\section{Dicussion and Future Work}
A proper evaluation metric in the literature of keyboard decoding should focus on leveraging contextual information while evaluating the performance of the decoder. Contextual information can provide improved correlations between various semantics of the sentence structure and the performance of the decoder, especially in word sense disambiguation. The Character Error Rate (CER) is not a smooth measure of the performance of the decoder. An appropriate evaluation method should evaluate semantics in multiple different contexts with respect to different subjects and intentions of the subjects. If the evaluation metric does not measure the performance of the model in a context aware perspective, the most high performing models will predict poorly when contexts and domain dynamically change. In fact, the loss function that the model is trained on should factor this contextual information and accordingly train the model. 

As shown in Tables~\ref{tab:context} and~\ref{tab:Googlecomp}, a model may be good at achieving a low CER in its predictions while it fails to capture the underlying context of the sentence. Google's search auto-corrects `yello world' to `yellow world' that is a syntactically correct prediction. However, our model predicts `hello world' which is more semantically and contextually correct. A model selected based on validation performance on the dev sets such that it has close CER to the ground truth but also generalizes well to different contexts under ambiguous usages, is a better model from true text processor perspective.

Therefore, the evaluation measure for model performance should factor a close coupling of the model between correction and contextual appropriateness. Performance evaluation in terms of CER alone will penalize models that can generalize better to contexts. For example, when the model prediction is `Don't they have some police there' for the ground truth `Don't they have some conflicts there', gives a high CER. However, the model seem to understand the context of the sentence, the connection between `police' and `conflicts', the appropriateness of `there' among others.

Therefore, here we attempt to propose an alternative evaluation measure that connects context to correction as an interesting future work. Consider there is an annotated Oracle corpus that has access to the context labels for each sentence in the corpus given by $S_c$, where $c$ is the context label. The context is a label that classifies the intent of the underlying sentence. The evaluation metric should measure if the model prediction has correctly captured the intent of the ground truth sentence and also penalize the model for very high CER values. Let the penalization term be given by $r$:
\begin{equation*}
r_c = \begin{cases}
1 \hspace{2.8cm}\text{if} \hspace{0.5cm} p_{cer} \leq {o_{cer}} \\
e^{\left(\frac{p_{cer}}{o_{cer}}\right)}e^{\left({1 - e^{P_{S_c}}}\right)} \hspace{0.5cm} \text{if} \hspace{0.5cm} p_{cer} > {o_{cer}}
\end{cases}
\end{equation*}
where, $p_{cer}$ is the CER of the model prediction with respect to the ground truth, $o_{cer}$ is the CER of the input stimulus with respect to the ground truth.
$P_{S_c}$ is the conditional probability of the context label $c \in C$ with respect to the word predicted by the model $w_t$ and the sequence $s_{t-1}$ seen at time $t$ such that $s_{t-1} = w_1, \dots, w_{t-1}$. If the CER of the model prediction is higher than the original CER, the model is penalized. By original CER we mean the CER between the input and the ground truth. We have,
\begin{equation*}
P_{S_c} = p(c_t|w_t,s_{t-1}).
\end{equation*}
Let the cross entropy of the model's prediction be given by:
\begin{equation*}
H_{S_c} = 	\frac{1}{N} \sum_{t=1}^{T}\log_{2}P_{S_c}.
\end{equation*}
The evaluation measure then captures the cross entropy of the model's prediction that accommodates the model to obtain credit for predicting a word $w_t$ with respect to the correct context $c_t$ even at the cost of high CER. It also allows for the model to not get credit for predicting $w_t$ for the incorrect context $c_t$ with low CER. Let this smoother version of the CER measure be denoted as $s_{CER}$ and is given by:
\begin{equation*}
s_{CER} = r_c \times H_{S_c}.
\end{equation*}
It will be interesting to see how our model and the state-of-the-art perform when evaluated using this smoother version of the CER measure.
\begin{table}
	\centering
	\begin{tabular}{l c c }
		\\
		\cmidrule(r){1-3}
		{\textit{Model}}
		& { \textit{Inputs}}
		& {\textit{Prediction}}\\
		\midrule
		\multirow{8}{*}{Google Search} 
		& 'I toks you silky' &' i told you silky' \\
		& ' lease stnd p' & ' please stand p' \\
		& 'sh ws s fnyy' & ' she was funny' \\ 
		& 'i hatttttte yyyoooouuuuu' & ' i hatttttte yyyooouuu' \\ 
		& 'tlk t y frnd' & 'talk t y frnd' \\
		& 'dont yello at me' & don't yell at me' \\ 
		& 'yello world' & 'yellow world' \\ 
		& ' i prefer yello tops' & 'i prefer yellow tops \\ 
		& 'ths i extrmley hrd' & 'this i extremely hard'\\
		& 'hes is rel slw' & 'hes is real slow'\\ 
		\midrule
		\multirow{8}{*}{CCEAD (Ours)} 
		& 'I toks you silky' &' i told you silly' \\
		& ' lease stnd p' & ' please stand up' \\
		& 'sh ws s fnyy' & ' she was so funny' \\ 
		& 'i hatttttte yyyoooouuuuu' & ' i hate you' \\ 
		& 'tlk t y frnd' & 'talk to my friend' \\
		& 'dont yello at me' & don't yell at me' \\ 
		& 'yello world' & 'hello world' \\ 
		& ' i prefer yello tops' & 'i prefer yellow tops \\ 
		& 'ths i extrmley hrd' & 'this is extremely hard'\\
		& 'hes is rel slw' & 'he's is real slow'\\ 
	\end{tabular}
	\caption{Our model predictions versus Google's search query corrections. }~\label{tab:Googlecomp}
\end{table}
\section{Conclusion}
In this paper, we propose a recurrent and convolutional neural sequence to sequence model for text correction and completion.  We show that our model achieves an average CER of $2.6$ on touch-screen keyboard and $2.1$ on physical keyboard. All traditional text correction and completion keyboard decoders train on billions of words and train over extensive period of time. In contrast,  our model trains on a small vocabulary of $50000$ words and the training is completed within a day with a model size an order of magnitude smaller than the conventional decoders. We perform validation over a synthetic dataset and two real world datasets. We also perform an in-house user study where users' typing data are collected on touch-screen keyboard and physical keyboard and evaluated. Further, we propose an alternative smoother evaluation measure over the character error rate (CER) for evaluating model predictions based on the underlying intent of the sentence.
\bibliographystyle{IEEEtran}
\bibliography{bib/Decoder-ref}
\end{document}